\documentclass[11pt]{article}

\usepackage[final]{acl}
\usepackage{times}
\usepackage{latexsym}

\usepackage[T1]{fontenc}

\usepackage[utf8]{inputenc}

\usepackage{microtype}

\usepackage{inconsolata}

\usepackage{graphicx}
\usepackage{booktabs}

\title{WAKMan: Workload Aware KVCache Manager}

\author{
  \textbf{Suhas Kumar Morisetty\textsuperscript{1,*}},
  \textbf{Vaibhavi Lokegaonkar\textsuperscript{2,*}},
  \textbf{Sachin Mathew Varghese\textsuperscript{3}},
  \textbf{Wesley Harris\textsuperscript{3}}
}

\begin{document}
\maketitle
{\let\thefootnote\relax\footnotetext{
  \textsuperscript{*}Equal contribution. Author order was determined by a coin toss. Work done during internship at Capital One. \\
  \textsuperscript{1}Columbia University, \texttt{vm2825@columbia.edu} \\
  \textsuperscript{2}University of Maryland, College Park, \texttt{vlokegao@umd.edu} \\
  \textsuperscript{3}Capital One, \texttt{\{sachinmathew.varghese, wesley.harris\}@capitalone.com}
}}
\begin{abstract}
Prefix caching enables a Large Language Model (LLM) servers to reuse the Key-Value (KV) states and accelerate prefill computation stage for requests that share a prefix with earlier prompts, thus reducing latency and serving cost. With limited memory resource, the KV cache eviction policy becomes a decisive factor in determining the overall benefits of prefix caching. Production traffic may have single-turn, multi-turn, agentic, or a mixture of any combination of these request types. Our findings indicate that agentic steps return within seconds and favor recency, while popular prompts and long conversations return after many other requests and favor frequency. Across 20 mixes, no single policy is always best: vLLM's default strategy LRU provides 12\% higher cache hits as compared to the frequency-oriented S3FIFO on agentic-heavy mixes but trails it by up to 22\% elsewhere, and even a single workload parameter can flip the winner even for any particular mix. We present WAKMan, a workload-aware KV cache manager that picks its eviction policy from live traffic without knowing request types or reading prompts. It improves hit rate over LRU by 5 percentage points on average, with the lowest mean and worst-case optimality gap of all policies.

\end{abstract}

\section{Introduction}
\begin{figure}
    \centering
    \includegraphics[width=0.75\linewidth]{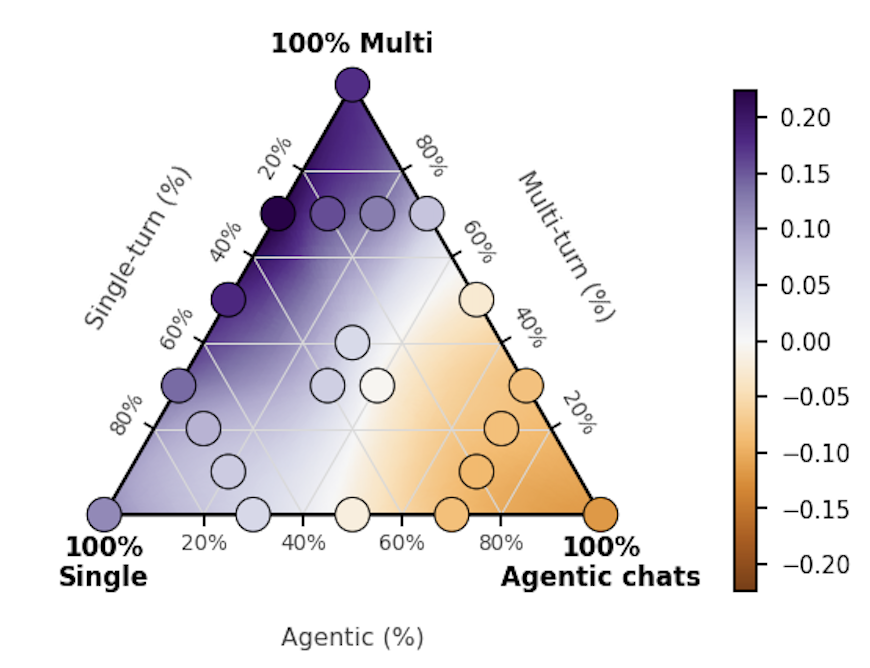}
    \caption{Hit-rate difference (S3FIFO $-$ LRU) across 21 request mixes. Purple: S3FIFO better. Yellow: LRU better. No single policy wins everywhere.}
    \label{fig:marquee}
    \vspace{-4pt}
\end{figure}

Serving Large Language Models (LLMs) efficiently at scale has become a prominent challenge in system design. LLM inference process involves generating the key and value vectors of each prompt token based on its preceding context. Requests that share a prompt prefix therefore produce identical KV states, and storing these can avoid unwanted recomputation. Most LLM servers store this KV cache in the GPU memory so that future requests with a common prefix can reuse these KV blocks. Since GPU memory is limited, cached blocks require eviction in order to store newer KV blocks. KV blocks will have to be recomputed if a particular block is required immediately after it gets evicted thus increasing the latency and computation cost for that request. Hence, selecting which KV blocks to evict is critical for extracting maximum performance from an LLM system, and the right strategy depends on how blocks are reused across requests.

\noindent LLM serving traffic can be a mix of three types of requests: single-turn, multi-turn, and agentic (Appendix~\ref{sec:workloads}). These differ in length of prefix they share and how quickly a shared prefix is requested again. vLLM\cite{kwon2023efficient} evicts blocks in Least Recently Used (LRU) order by default. LRU strategy works well when shared prefixes return quickly, as in agentic sessions. It fares sub-optimally when many one-time blocks, such as single-turn payloads, push out blocks that will be reused later, such as fixed prompts and long conversation histories. Scan-resistant policies such as S3FIFO protect these blocks, but they can evict fast-returning agentic blocks before their next use. Figure~\ref{fig:marquee} shows the this effect across mixes of the three request types. S3FIFO is better when single-turn or multi-turn requests dominate, while LRU is better when agentic requests dominate. The better policy also changes when the mix is fixed and a workload parameter, such as the number of turns per conversation, changes (Figure~\ref{fig:ablation}). Since traffic also shifts over time, this makes selecting a policy upfront suboptimal.


\noindent Recent systems improve on LRU strategy by making KV cache eviction aware of the request type. UniCache~\citep{yang2026unicachetasktypeawarekv} and SAECache~\citep{fang2026tokensworthcachinglearning} keep separate queues for agentic, chat, and single-turn requests, and a study of Alibaba's production traces models KV cache reuse by request type and turn number~\citep{wang2026kvcachecachewildcharacterizing}. Other systems predict when a prefix will be reused, either with a trained model~\citep{fang2026tokensworthcachinglearning} or with inter-access time distributions fit on past traces~\citep{yang2026unicachetasktypeawarekv, wang2026kvcachecachewildcharacterizing, li2026continuumefficientrobustmultiturn}. These methods make two assumptions: the server knows each request's type and turn, and reuse timing stays stable over time. Neither holds in any large-scale shared deployment serving multiple use cases. As requests arrive, conversation or agent history is concatenated on the client side. The server cannot recover a request's type or turn due to data-sharing and data-privacy restrictions in most shared deployments. The traffic mix also keeps shifting, so a distribution fit on past traces may no longer hold. A policy for serving must therefore rely only on what the cache itself observes.



\noindent We propose WAKMan, a workload-aware KV cache manager that selects its eviction policy from live traffic (Figure 1). WAKMan runs shadow copies of LRU and S3FIFO alongside the server. At fixed intervals, it replays the most recent trace on both and switches the server to the policy with the higher simulated hit rate. WAKMan uses only the block hashes that vLLM already computes. It needs no request types, no prompt content, and no offline training, and it does not change model outputs. Unlike adaptive policies such as ARC~\cite{270366} and LeCaR~\cite{vietri2018lecar}, which adjust within a single cache on every request, WAKMan selects one complete policy per interval (Section \ref{sec:method}). This keeps the eviction path simple, and any policy that can be simulated can join the candidate pool.

\noindent We evaluate WAKMan on 19 synthetic workloads with different mixes of single-turn, multi-turn, and agentic requests, served with Gemma on vLLM. WAKMan improves the hit rate over LRU, the vLLM default, by 5 points on average. To test whether a single candidate policy would suffice, we also run S3FIFO n. S3FIFO is better than LRU on most workload mixes, but it falls well behind on agentic-heavy mixes, where LRU is better. WAKMan follows the policy that fits the current traffic, and it has the lowest gap between its cache hit rate and that of the ideal cache policy of the three policies.


\vspace{-3pt}
\section{Methodology}
\label{sec:method}
\begin{figure*}[t]
    \centering
    \includegraphics[width=0.85\linewidth]{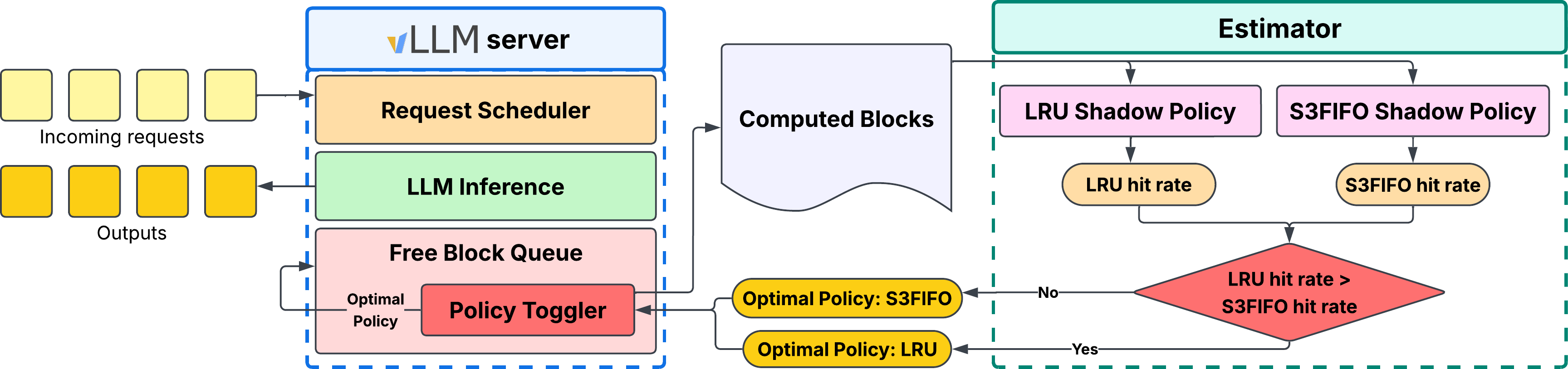}
    \caption{\textbf{Overview of WAKMan.} Every $T$ minutes, the estimator replays recent block hashes through shadow LRU and S3FIFO policies, and the server switches to the policy with the higher hit rate.}
    \label{fig: dynamic policy}
\vspace{-5mm}
\end{figure*}
Our method switches the eviction policy of vLLM~\cite{kwon2023efficient} between LRU and S3FIFO~\cite{10.1145/3600006.3613147} at run time. Every $T$ minutes, it asks which of the two policies would have served the last $T$ minutes of traffic better, and uses that policy for the next $T$ minutes (Figure~\ref{fig: dynamic policy}). We initially seed vLLM with LRU. During each window of $T$ minutes, we log the arrival time and the block hashes of every request. We also track the average size of the free block queue over the window.

\noindent\textbf{Estimator.} At the end of each window, the estimator replays the logged requests in arrival order through two shadow policies, one for LRU and one for S3FIFO. The estimator mimics the eviction of vLLM as closely as possible without a GPU or a model runner, using only the block hashes from vLLM. Both shadow policies get a cache capacity equal to the average size of the free block queue, since this is roughly the space that the eviction policy manages. Below, we describe the logic of LRU \& S3FIFO used in both the vLLM implementation and the estimator.

\noindent\textbf{Shadow LRU policy.} The shadow LRU policy maintains a single queue of block hashes ordered by recency. A hit is counted when an incoming block hash is already present in the queue. On a miss, the block is inserted and the least recently used block is evicted if the cache is full. The hit rate $hit_{LRU}$ is the number of hits divided by the total number of block hashes seen in the last $T$ minutes.

\noindent\textbf{Shadow S3FIFO policy.} The shadow S3FIFO policy keeps small, main, and ghost queues. Unlike the original, the small and main queues resize freely, with their total equal to the available GPU blocks. New blocks enter the small queue and move to the main queue if hit. Otherwise, they go to the ghost queue, and returning ghost blocks enter the main queue directly. The FIFO main queue reinserts blocks with a nonzero 2-bit hit counter rather than evicting them. $hit_{S3FIFO}$ is computed as for LRU.

\noindent\textbf{Switching criteria.} Every $T$ minutes, both shadow policies are replayed on the trace from the last $T$ minutes. If $hit_{S3FIFO} > hit_{LRU}$ \footnote{$hit_{LRU}$ is hit rate from running shadow LRU policy}, the server uses S3FIFO for the next $T$ minutes. Otherwise, it uses LRU. When the server switches from S3FIFO to LRU, the ghost queue entries are discarded.
\vspace{-3pt}

\section{Experiments}
\label{sec:experiments}
\vspace{-3pt}
\subsection{Workload Data}
\label{sec:data}

We use only synthetic workloads. Synthetic traces let us set the properties of each request type and change one property at a time.  Each trace is one hour long, with an average arrival rate of 1.5 requests per second.\footnote{We tuned this rate to put the cache under pressure. At lower rates, the cache rarely evicts. At higher rates, blocks are evicted before any reuse, so the eviction policy does not matter.} We sample each prompt, input, and output length from a normal or log-normal distribution with a set mean or median. Appendix~\ref{sec:appendix} gives the full configuration.

\noindent\textbf{Single-turn requests.}
Single-turn requests draw their fixed prompts from a catalog of 100 services, with prompt lengths from roughly 300 to 8K tokens. We model the skewed popularity from Section~\ref{sec:workloads} with a Zipf--Mandelbrot law, so the request rate of a service decreases with its popularity rank. We set roughly 15\% of requests have no fixed prompt. 

\noindent\textbf{Multi-turn requests.}
We model both the number of turns per session and the inter-turn time with log-normal distributions, with medians of three turns and 60 seconds.

\noindent\textbf{Agentic requests.}
We model the number of turns per session and the inter-turn time with log-normal distributions. A session has 5 to 15 turns, with a median of 7. The inter-turn time has a median of 6 seconds and a cap of 300 seconds. Most turns add a short tool result, but 15\% add a long context read of about 3K tokens.

\noindent\textbf{Heterogeneous sweep.}
We generate 20 workloads that vary only the mix of single-turn, multi-turn, and agentic sessions, from pure types to equal thirds, with all other parameters fixed. We write a mix as S/M/A, the percentage of each type.
\vspace{-3pt}
\subsection{Setup}
\label{sec:setup}

We serve \texttt{google/gemma-4-26B-A4B} in bf16 on vLLM with tensor parallelism across four NVIDIA RTX PRO 6000 Blackwell GPUs (96\,GB each), with GPU memory utilization 0.5 and a block size of 16 tokens. LRU, S3FIFO, and WAKMan run inside vLLM as Python implementations. WAKMan starts with LRU and decides every $T=15$ minutes. AIPerf replays each trace at its exact timestamps. We report the prefix-cache hit rate, the fraction of prompt tokens read from the KV cache. The ideal hit rate uses an infinite cache and bounds any policy. A policy's regret on a workload is the best hit rate among all policies minus its own.
\vspace{-3pt}
\subsection{Results}
\label{sec:results}

\begin{figure*}[t]
\centering
\includegraphics[width=0.85\linewidth]{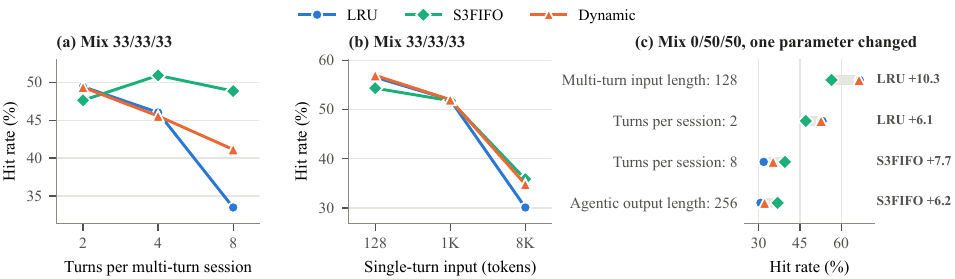}
\caption{The best static policy changes within a fixed request mix. (a, b) The 33/33/33 mix as one parameter grows. (c) Variants of the 0/50/50 mix, each with one parameter changed. The gray bar spans LRU and S3FIFO, and the label gives the better static policy and its margin in points.}
\vspace{-4pt}
\label{fig:ablation}
\end{figure*}

\begin{table}[t]
\centering
\small
\begin{tabular}{@{}lrrrr@{}}
\toprule
Mix S/M/A & LRU & S3FIFO & Dynamic & Ideal \\
\midrule
100/0/0 & 37.8 & \textbf{49.5} & 46.5 & 50.1 \\
70/0/30 & 55.3 & \textbf{60.0} & 57.4 & 70.0 \\
50/0/50 & \textbf{61.3} & 59.5 & \textbf{61.3} & 76.1 \\
30/0/70 & \textbf{64.4} & 56.1 & 64.3 & 80.3 \\
0/0/100 & \textbf{69.3} & 57.6 & 69.2 & 84.7 \\
70/10/20 & 49.8 & \textbf{55.6} & 54.0 & 66.1 \\
20/10/70 & \textbf{62.4} & 53.4 & 62.3 & 80.3 \\
70/20/10 & 43.5 & \textbf{51.7} & 49.4 & 62.3 \\
10/20/70 & 60.4 & 52.1 & \textbf{60.4} & 80.9 \\
70/30/0 & 31.5 & \textbf{45.4} & 42.4 & 56.9 \\
30/30/40 & 50.9 & 50.4 & \textbf{51.1} & 74.3 \\
30/40/30 & 42.8 & \textbf{47.3} & 42.8 & 73.2 \\
0/30/70 & \textbf{59.1} & 51.0 & \textbf{59.1} & 81.4 \\
50/50/0 & 23.4 & \textbf{41.4} & 38.6 & 60.4 \\
0/50/50 & \textbf{49.5} & 46.8 & 49.2 & 78.6 \\
30/70/0 & 14.2 & \textbf{36.6} & 34.1 & 63.1 \\
20/70/10 & 22.6 & \textbf{38.3} & 36.1 & 67.6 \\
10/70/20 & 27.4 & \textbf{39.9} & 37.9 & 71.7 \\
0/70/30 & 34.3 & \textbf{40.9} & 38.3 & 75.2 \\
0/100/0 & 8.4 & \textbf{26.0} & 15.0 & 67.1 \\
\midrule
Mean regret & 7.09 & 2.53 & \textbf{2.04} & \\
Max regret & 22.4 & 11.7 & \textbf{11.0} & \\
\bottomrule
\end{tabular}
\caption{Prefix-cache hit rate (\%) on the 20 workload mixes. Bold shows the best policy in each row. Ideal is the hit rate with an infinite cache.}
\vspace{-10pt}
\label{tab:gold}
\end{table}
\vspace{-3pt}
Table~\ref{tab:gold} gives the hit rates on the 20 mixes. No static policy is the best on all mixes. Between the two static policies, LRU wins on 8 mixes and S3FIFO wins on 12. Each static policy also has a large loss on some mixes. LRU loses up to 22.4\% of cache hits, and S3FIFO loses up to 11.7\% of the same.

\noindent\textbf{Agentic workloads favor LRU.}
Agentic turns arrive quickly, so the next turn of a session usually finds its history at the recent end of the cache. Thus, recency is a good signal for reuse, making LRU the better policy. S3FIFO puts each new block in its small queue first and evicts agentic blocks from it before the next turn arrives.

\noindent\textbf{Single-turn workloads favor S3FIFO.}
Single-turn reuse depends on frequency. The fixed prompts of popular services get many hits, but the payloads get no hits. S3FIFO keeps the hot fixed prompts in its main queue and removes each payload quickly from the small queue. Thus, S3FIFO protects the fixed prompts from scan pollution by the payloads, making its hit rate close to ideal. 

\noindent\textbf{Multi-turn workloads favor S3FIFO.}
In a multi-turn workload, many other requests arrive during the long gap between two turns. These requests push the conversation history to the old end of the LRU queue, and LRU evicts it. But each turn reads the full history again, so the blocks of earlier turns get more hits as the conversation continues. S3FIFO moves these blocks to its main queue and keeps them across the gap. 

\noindent\textbf{The dynamic policy is the most robust.}
The dynamic policy has the lowest mean and maximum regret of the three policies. When LRU and S3FIFO differ by more than one point, the dynamic policy recovers a median of 84\% of the difference.The largest regret is on the pure multi-turn mix (0/100/0) where the dynamic policy suffers due to its initialization with LRU. The dynamic policy starts with LRU and makes its first switch only after 15 minutes, a quarter of each one-hour trace. On longer traces, this startup cost shrinks.
\subsection{Ablations}
\label{sec:ablations}

The main sweep shows that the best policy depends on the request mix. The ablations aim to seek answers to two questions. First, does the best policy also change when the mix stays fixed and one workload parameter changes? Second, if it changes, does the dynamic policy adapt? 

\noindent We use the balanced 33/33/33 mix and change either the turns per multi-turn session or the single-turn input length. We also change single parameters of the 0/50/50 mix.

\noindent\textbf{The mix alone does not determine the best policy.}
Figure~\ref{fig:ablation} shows that a single parameter, such as number of turns in multi-turn requests, can flip the better static policy. On 33/33/33, LRU wins with 2 turns per session, but S3FIFO wins by 15.4 points with 8 turns. Single-turn input length has the same effect, favoring LRU at 128 tokens and S3FIFO at 8K tokens. On 0/50/50, either policy can win by up to 10 points depending on the parameter. Since these parameters vary with users and applications, an operator cannot fix one policy in advance.

\noindent\textbf{The dynamic policy adapts.}
The dynamic policy uses neither the mix nor the workload parameters. On every point in Figure~\ref{fig:ablation}, it never falls more than 0.5 points below the worse static policy. Where LRU is better, the dynamic policy matches LRU. Where S3FIFO is better, it moves toward S3FIFO.
\vspace{-3pt}
\section{Conclusion}
\label{sec:conclusion}

We studied KV cache eviction in LLM serving under heterogeneous workloads. vLLM's default LRU policy is not the best for all workloads, since the better policy changes with the request mix and with single workload parameters. WAKMan selects the better policy at run time by replaying recent traffic, and it has the lowest mean regret of the three policies.

\section*{Limitations}

We use only LRU and S3FIFO~\citep{10.1145/3600006.3613147} as candidate policies. Any policy that a simulator can replay, such as ARC~\citep{270366}, can join the pool and add headroom where both candidates are far from the ideal hit rate. Our goal was not to find the best eviction policy, but to show that trace replay can help in selecting a better policy at run time.


\noindent Determining the exact or exploring the need for a dynamically changing time interval T to provide the maximum possible cache hit rates needs further analysis. 




\bibliography{custom}

\appendix

\section{Potential risks} 

WAKMan changes only which KV blocks stay in GPU memory. It does not change model outputs, so it adds no risk to the content, safety, or fairness of generated text. We see three operational risks. First, a switch can be wrong for an entire interval. WAKMan chooses the policy for the next $T$ minutes based on the traffic of the previous $T$ minutes. If the traffic changes sharply right after a decision, the server keeps a mismatched policy until the next decision, and requests in that interval see a lower hit rate, higher latency, and higher cost. Second, prefix caching can leak information through timing. A request whose prefix is already cached skips prefill and receives its first token sooner. A user who shares the server can send a guessed prefix and measure this delay to learn whether another user recently sent the same prefix, such as a particular system prompt or document. WAKMan does not create this channel, which exists for any prefix cache. However, it changes which prefixes stay cached and for how long. For example, S3FIFO keeps popular prefixes longer than LRU, which can keep the signal observable for longer. WAKMan should therefore be deployed with the isolation that operators already use for prefix caching, such as separate caches for each user or tenant. Third, switching has its own cost. The estimator replays the traffic of every interval, and each switch changes how the cached blocks are ordered for eviction. For some traffic patterns, this overhead can exceed the benefit of the extra cache hits. This can happen when the two policies have similar hit rates, so a switch gains little, or when traffic alternates quickly between recency-heavy and scan-heavy phases, so one switch is soon followed by another. In such cases, a static policy may be cheaper overall.

\section{License of artifacts}
We use vLLM~\cite{kwon2023efficient}, the Gemma~4 model (\texttt{google/gemma-4-26B-A4B}), and AIPerf. All three are released under the Apache License 2.0, which permits research and commercial use, modification, and redistribution. Our use of each artifact is consistent with its license and intended use. All workloads in this paper are synthetic and generated by our own trace generator, so they contain no personal or user data. 


\section{Use of AI Assistants}
We used Claude, an AI assistant, to help with writing. It helped restructure and shorten drafts of the introduction and methodology, suggest phrasing for figure captions, and draft parts of the Limitations section. All research ideas, system design, implementation, experiments, and results are our own. The authors reviewed and edited all AI-assisted text and take full responsibility for the content of the paper.

\section{Background}
KV cache optimizations fall into three families. Offloading moves key-value state across the GPU, CPU, and disk hierarchy. Compression shrinks the cache through quantization, low-rank approximation, or token pruning . Exact eviction keeps full-precision blocks on the GPU and only chooses which to drop under memory pressure. Our work is an exact-eviction method. \\
\subsection{KV Cache Compression Strategies}
\label{sec:related-compression}

Quantization stores keys and values with fewer bits per entry, through rotation-based vector quantization~\cite{zuo2026ratequantoptimalmixedprecisionkv}, layer-wise and head-wise mixed precision~\cite{li2025kvtunersensitivityawarelayerwisemixedprecision,zuo2026ratequantoptimalmixedprecisionkv}, or hardware co-design with offline-profiled outlier thresholds~\cite{Kim_2025}. Transform coding and low-rank methods remove redundancy in the feature dimension. Palu~\cite{chang2024palucompressingkvcachelowrank} and xKV~\cite{chang2026xkvcrosslayerkvcachecompression} use within-layer and cross-layer low-rank factorization, KVTC~\cite{staniszewski2026kvcachetransformcoding} adds PCA, adaptive quantization, and entropy coding for stored caches, and ShadowServe~\cite{xiang2025shadowserveinterferencefreekvcache} offloads decompression to a SmartNIC. Latent compaction shortens the sequence by fitting a smaller cache instead of keeping a subset of the original. Attention Matching~\cite{zweiger2026fastkvcompactionattention} fits compact keys and values to match full-cache attention, and Liu et al.~\cite{liu2026practicalonlinekvcache} adapt it to online agents. Token selection keeps the KV entries of important tokens and evicts the rest. CompressKV~\cite{lin2026compresskvsemanticretrievalguidedkvcachecompression} scores tokens with semantic retrieval heads, and KVzip~\cite{kim2025kvzipqueryagnostickvcache} scores tokens by how well they reconstruct the context. These methods compress the KV state of a single request and mostly fix their choices offline. Our work is complementary: it decides which cached blocks to keep across requests and adapts the policy online. Unlike the compression methods above, our method does not alter the cached KV blocks. It keeps each block at full precision and instead decides which blocks to retain and which to evict. Our method therefore belongs to the family of eviction-based techniques, and it can be combined with compression for added benefits to cache hit rates. 
\subsection{KV Cache Eviction Strategies}
\label{sec:related-eviction}
Eviction-based methods keep cached KV blocks intact and decide which blocks to retain when GPU memory runs out. Production engines such as vLLM and SGLang default to LRU~\cite{xie2026peekpredictivequeueinformedkv}. Recency and frequency policies refine this default. Liu et al.~\cite{liu2026fancyevictionfailsrethinking} evaluate 14 policies on production traces and find that recency is unusually predictive, so sophisticated policies gain little over LRU. ARC-based caching~\cite{shen2026recencyfrequencyadaptivekvcaching} adapts the split between recency and frequency queues using ghost caches, and UniCache~\cite{yang2026unicachetasktypeawarekv} captures multiple reuse patterns and balances cache allocation across tasks in heterogeneous workloads. Workload-aware and learned policies predict reuse directly. Wang et al.~\cite{wang2026kvcachecachewildcharacterizing} use per-workload reuse probability distributions, LPC~\cite{NEURIPS2025_414f642a} predicts which conversations will continue, and SAECache~\cite{fang2026tokensworthcachinglearning} learns the reuse value of each token type online. Cost-aware policies weigh the price of a miss. Marconi~\cite{pan2024marconi} uses FLOP efficiency for hybrid models, and EvicPress~\cite{feng2025evicpressjointkvcachecompression} jointly decides compression and eviction across storage tiers. Agent-aware policies exploit program structure. KVFlow~\cite{pan2025kvflowefficientprefixcaching} ranks agents by steps to execution, Continuum~\cite{li2026continuumefficientrobustmultiturn} pins KV blocks during tool calls with a time-to-live, PBKV~\cite{zheng2026efficientservingdynamicagent} predicts agent invocations in dynamic workflows, and RR-Evict~\cite{pan2026rrevictfinegrainedprefixcache} spreads reclamation across idle agents. PEEK~\cite{xie2026peekpredictivequeueinformedkv} uses the waiting queue to protect prefixes that pending requests will need. Most of these methods commit to a single policy or tune one policy online. In contrast, our method selects between whole policies at run time based on their simulated hit rate on recent traces.

\section{Types of Workloads}
\label{sec:workloads}

A production LLM service receives requests from many applications through one endpoint. We group these requests into three types: single-turn, multi-turn, and agentic. The three types differ in two properties that control the reuse of KV cache blocks, namely the shared prefix and the inter-turn time.

\noindent\textbf{Single-turn workloads.}
A single-turn workload consists of independent requests, such as API calls from services or sandboxes, each answered in one turn. Each request combines a fixed prompt, a template reused across many requests, with a unique payload. Popularity is skewed: a few fixed prompts receive most requests. Requests with the same fixed prompt therefore share it as a prefix, but no payload is ever reused.

\noindent\textbf{Multi-turn workloads.}
A multi-turn workload consists of conversations between users and the LLM, as in a chatbot. Each conversation starts with a context specific to its user, such as a system prompt combined with user memory or reference documents. Each turn appends the previous response and a new message to the history, so the prefix grows and the KV cache can supply most of the inpu. But the inter-turn time is long: the user reads, thinks, and types, which takes seconds to minutes.

\noindent\textbf{Agentic workloads.}
An agentic workload consists of automated workflows in which the LLM takes a sequence of steps. Each step appends the result of the previous step, such as a tool output, to the workflow history, so the prefix grows as in a conversation. Agentic sessions run more turns than conversations, and with no human in the loop, the inter-turn time is short.\footnote{Long tool calls or code execution can lengthen the gap between agent steps. Our heterogeneous workloads cover long gaps through their multi-turn fraction.}

\noindent\textbf{Heterogeneous workloads.}
A shared deployment may receive all three types at once, with their requests interleaved. The proportions also change over time, for example when chat traffic peaks during work hours and batch jobs run overnight. The mix determines which KV cache eviction policy performs best, and in this paper we analyze how this choice changes across proportions.

\section{Workload Configuration}
\label{sec:appendix}

This configuration generates the main workloads in Section~\ref{sec:data}. For each of the 20 mixes, we change only the \texttt{frac} value of each profile. The \texttt{frac} values give the fraction of new sessions of each type. In each \texttt{system\_prompts} and \texttt{mixture} list, the generator draws a length from a log-normal distribution with the given median and log-space spread (\texttt{sigma}). The \texttt{catalog} gives the number of single-turn services and the parameters of their Zipf--Mandelbrot popularity, $p(r) \propto (r+q)^{-\alpha}$.

\begin{small}
\begin{verbatim}
seed: 0
duration_s: 3600
arrival_rate: 1.5
profiles:
  single:
    frac: 1.0
    share_head: true
    no_system_prompt_frac: 0.15
    catalog: {n_services: 100, alpha: 1.3, q: 5}
    system_prompts:
    - {weight: 0.1, median: 8000, sigma: 0.05}
    - {weight: 0.6, median: 2500, sigma: 0.2}
    - {weight: 0.2, median: 1000, sigma: 0.2}
    - {weight: 0.1, median: 300,  sigma: 0.25}
    input:  {mean: 2048, std: 128, skew: right}
    output: {mean: 512,  std: 256, skew: left}
  multi_turn:
    frac: 0.0
    share_head: false
    min_turns: 2
    turns: {median: 3, sigma: 0.6}
    inter_turn_time: {median: 60.0, sigma: 0.5}
    system_prompts:
    - {weight: 0.2, median: 1250, sigma: 0.2}
    - {weight: 0.5, median: 500,  sigma: 0.2}
    - {weight: 0.3, median: 150,  sigma: 0.25}
    input:  {mean: 1024, std: 128}
    output: {mean: 768,  std: 64}
  agentic:
    frac: 0.0
    share_head: false
    min_turns: 5
    max_turns: 15
    turns: {median: 7, sigma: 1.7}
    inter_turn_time: {median: 6, sigma: 2.5}
    max_delay_s: 300
    system_prompts:
    - {weight: 0.4, median: 1024, sigma: 0.2}
    - {weight: 0.4, median: 2048, sigma: 0.2}
    input:
      mixture:
      - {weight: 0.85, median: 600,  sigma: 0.5}
      - {weight: 0.15, median: 3000, sigma: 0.2}
    output: {median: 126, sigma: 1.4}
\end{verbatim}
\end{small}

\end{document}